\documentclass{article}
\usepackage{amsmath,amsfonts,amsthm,mdframed,graphicx,subcaption,hyperref}
\usepackage{iclr2016_conference,times}

\DeclareMathOperator*{\argmin}{arg\,min}
\newcommand{\softmax}{\sigma}

\title{Unifying Distillation\\and Privileged Information}
\author{David Lopez-Paz\\
  Facebook AI Research, Paris, France\thanks{The majority of this work was done
  while DLP was affiliated to the Max Planck Institute for Intelligent
  Systems and the University of Cambridge.}\\
  \texttt{dlp@fb.com}
  \AND
  L\'eon Bottou\\
  Facebook AI Research, New York, USA\\
  \texttt{leon@bottou.org}
  \AND
  Bernhard Sch\"olkopf\\
  Max Planck Insitute for Intelligent Systems, T\"ubingen, Germany\\
  \texttt{bs@tuebingen.mpg.de}
  \AND
  Vladimir Vapnik\\
  Facebook AI Research and Columbia University, New York, USA\\
  \texttt{vladimir.vapnik@gmail.com}
}

\iclrfinalcopy

\begin{document}

\maketitle

\begin{abstract}
\emph{Distillation} \citep{Hinton15} and \emph{privileged information}
\citep{VapIzm15} are two techniques that enable machines to learn
from other machines. 
This paper unifies the two into
\emph{generalized distillation}, a framework to learn from multiple machines
and data representations.
We provide theoretical and causal insight about the inner workings
of generalized distillation, extend it to unsupervised, semisupervised and
multitask learning scenarios, and illustrate its efficacy on a variety of
numerical simulations on both synthetic and real-world data.
\end{abstract}

\section{Introduction}\label{sec:intro}
Humans learn much faster than machines.
\citet{VapIzm15} illustrate this discrepancy with the Japanese proverb
\begin{center}
\emph{better than a thousand days of diligent study is one day with a great
teacher}.
\end{center}
Motivated by this insight, the authors incorporate an
``intelligent teacher'' into machine learning. Their solution 
is to consider training data formed by a collection of triplets
\begin{equation*}
 \{(x_1, x^\star_1, y_1), \ldots, (x_n, x^\star_n, y_n) \} \sim P^n(x,x^\star,y).
\end{equation*}
Here, each $(x_i, y_i)$ is a feature-label pair, and the novel element
$x^\star_i$ is additional information about the example $(x_i, y_i)$ provided
by an intelligent teacher, such as to support the learning process.
Unfortunately, the learning machine will not have access to the teacher
explanations $x^\star_i$ at test time. Thus, the framework of \emph{learning
using privileged information} \citep{Vapnik09,VapIzm15} studies how to leverage
these explanations $x^\star_i$ at training time, to build a classifier for test
time that outperforms those built on the regular features $x_i$
alone. As an example, $x_i$ could be the image of a biopsy, $x^\star_i$ the
medical report of an oncologist when inspecting the image, and $y_i$ a binary
label indicating whether the tissue shown in the image is cancerous or healthy.

The previous exposition finds a mathematical justification in VC theory
\citep{Vapnik98}, which characterizes the speed at which machines learn using
two ingredients: the capacity or flexibility of the machine, and the amount of
data that we use to train it. Consider a binary classifier $f$ belonging to a
function class $\mathcal{F}$ with finite VC-Dimension
$|\mathcal{F}|_\textrm{VC}$. Then, with probability $1-\delta$, the
\emph{expected error} $R(f)$ is upper bounded by
\begin{equation}
  R(f) \leq R_n(f) + O\left(\left(\frac{|\mathcal{F}|_{\textrm{VC}}-\log
  \delta}{n}\right)^{\alpha}\right),
\end{equation}
where $R_n(f)$ is the training error over $n$ data, and $\frac{1}{2} \leq
\alpha \leq 1$.  For difficult (\emph{non-separable}) problems the exponent is
$\alpha = \frac{1}{2}$, which translates into machines learning at a
\emph{slow} rate of $O(n^{-1/2})$.  On the other hand, for easy
(\emph{separable}) problems, i.e., those on which the machine $f$ makes 
no training errors, the exponent is $\alpha = 1$, which translates into machines
learning at a \emph{fast} rate of $O(n^{-1})$.  The difference between these
two rates is huge: the $O(n^{-1})$ learning rate potentially only requires
$1000$ examples to achieve the accuracy for which the $O(n^{-1/2})$ learning
rate needs $10^6$ examples.  So, given a student who learns from a fixed amount
of data $n$ and a function class $\mathcal{F}$, a good teacher can try to ease
the problem at hand by accelerating the learning rate from $O(n^{-1/2})$ to
$O(n^{-1})$.

Vapnik's \emph{learning using privileged information} is one example of what
we call \emph{machines-teaching-machines}: the paradigm where machines learn
from other machines, in addition to training data. Another seemingly unrelated
example is \emph{distillation} \citep{Hinton15},\footnote{Distillation 
relates to \emph{model compression} \citep{Burges97,Bucilua06,Ba14}. We will
adopt the term \emph{distillation} throughout this manuscript.} where a simple
machine learns a complex task by imitating the solution of a flexible machine.
In a wider context, the machines-teaching-machines paradigm is one step toward
the definition of \emph{machine reasoning} of \citet{bottou2014machine},
``the algebraic manipulation of previously acquired knowledge to
answer a new question''. In fact, many recent state-of-the-art systems compose
data and supervision from multiple sources, such as object recognizers reusing
convolutional neural network features \citep{oquab2014learning}, and natural
language processing systems operating on vector word representations extracted
from unsupervised text corpora \citep{mikolov2013efficient}.

In the following, we frame Hinton's distillation and Vapnik's privileged
information as two instances of the same machines-teaching-machines paradigm,
termed \emph{generalized distillation}. The analysis of generalized
distillation sheds light to applications in semi-supervised learning,
domain adaptation, transfer learning, Universum learning \citep{Weston06},
reinforcement learning, and curriculum learning \citep{bengio2009curriculum};
some of them discussed in our numerical simulations.

\section{Distillation}\label{sec:distillation}
  We focus on $c$-class classification, although the same ideas apply to
  regression. Consider the data
  \begin{equation}
  \label{eq:data1}
    \{(x_i, y_i)\}_{i=1}^n \sim P^n(x,y), \,\, x_i \in \mathbb{R}^d, \,\, y_i
  \in \Delta^c.
  \end{equation}
  Here, $\Delta^c$ is the set of $c$-dimensional probability vectors.  Using
  \eqref{eq:data1}, we are interested in learning the representation 
  \begin{equation}\label{eq:obj1}
    f_t = \argmin_{f \in \mathcal{F}_t} \frac{1}{n} \sum_{i=1}^n \ell(y_i, \softmax(f(x_i))) + \Omega(\|f\|),
  \end{equation}
  where $\mathcal{F}_t$ is a class of functions from $\mathbb{R}^d$ to
  $\mathbb{R}^c$, the function $\softmax : \mathbb{R}^c \to \Delta^c$ is the
  softmax operation
  \begin{equation*}
  \softmax(z)_k = \frac{e^{z_k}}{\sum_{j=1}^c e^{z_j}},
  \end{equation*}
  for all $1 \leq k \leq c$, the function $\ell : \Delta^c \times \Delta^c \to \mathbb{R}_+$
    is the cross-entropy loss
  \begin{equation*}
    \ell(y,\hat{y}) = -\sum_{k=1}^c y_k \log \hat{y}_k,
  \end{equation*}
  and $\Omega : \mathbb{R} \to \mathbb{R}$ is an increasing function which serves as a regularizer.

  When learning from real world data such as high-resolution images, $f_t$ is
  often an ensemble of large deep convolutional neural networks
  \citep{Lecun98}.  The computational cost of predicting new examples at test
  time using these ensembles is often prohibitive for production systems.  For
  this reason, \citet{Hinton15} propose to \emph{distill} the
  learned representation $f_t \in \mathcal{F}_t$ into
  \begin{equation}\label{eq:obj2}
    f_s = \argmin_{f\in \mathcal{F}_s} \frac{1}{n} \sum_{i=1}^n \Big[(1-\lambda)\ell(y_i,
    \softmax({f}(x_i))) + \lambda \ell(s_i,
    \softmax({f}(x_i)))\Big],
  \end{equation}
  where
  \begin{equation}\label{eq:soft}
    s_i = \softmax(f_t(x_i)/T) \in \Delta^c
  \end{equation}   
  are the \emph{soft predictions} from $f_t$ about the training data, and
  $\mathcal{F}_s$ is a function class  simpler than $\mathcal{F}_t$. The
  temperature parameter $T > 0$ controls how much do we want to soften or
  smooth the class-probability predictions from $f_t$, and the imitation
  parameter $\lambda \in [0,1]$ balances the importance between imitating the
  soft predictions $s_i$ and predicting the true hard labels $y_i$. Higher
  temperatures lead to softer class-probability predictions $s_i$. In turn,
  softer class-probability predictions reveal label dependencies which would be
  otherwise hidden as extremely large or small numbers. After distillation, we
  can use the simpler ${f_s} \in \mathcal{F}_s$ for faster prediction at test
  time.
  
\section{Vapnik's privileged information}
We now turn back to Vapnik's problem of learning in the company of an
intelligent teacher, as introduced in Section~\ref{sec:intro}. The question at
hand is: How can we leverage the privileged information $x^\star_i$ to build a
better classifier for test time?  One na\"ive way to proceed would be to estimate the
privileged representation $x^\star_i$ from the regular representation $x_i$, and then use
the union of regular and \emph{estimated} privileged representations as our
test-time feature space.  But this may be a cumbersome endeavour: in the
example of biopsy images $x_i$ and medical reports $x^\star_i$, it is
reasonable to believe that predicting reports from images is more
complicated than classifying the images into cancerous or healthy.

Alternatively, we propose to use distillation to extract useful knowledge from
privileged information.  The proposal is as follows. First, learn a teacher
function $f_t \in \mathcal{F}_t$ by solving \eqref{eq:obj1} using the data
$\{(x^\star_i, y_i)\}_{i=1}^n$.  Second, compute the teacher soft labels $s_i =
\sigma(f_t(x^\star_i)/T)$, for all $1\leq i\leq n$ and some temperature
parameter $T > 0$.  Third, distill $f_t \in \mathcal{F}_t$ into $f_s \in
\mathcal{F}_s$ by solving \eqref{eq:obj2} using both the hard labeled data
$\{(x_i, y_i)\}_{i=1}^n$ and the softly labeled data $\{(x_i,s_i)\}_{i=1}^n$.

\subsection{Comparison to prior work}

\citet{Vapnik09,VapIzm15} offer two strategies to learn using privileged information:
similarity control and knowledge transfer. Let us briefly compare them to 
our distillation-based proposal.

The motivation behind \emph{similarity control} is that SVM
classification is separable after we correct for the \emph{slack values}
$\xi_i$, which measure the degree of misclassification of training data points $x_i$
\citep{Vapnik09}.
Since separable classification admits
$O(n^{-1})$ fast learning rates, it would be ideal to have a teacher that could
supply slack values to us.  Unluckily, it seems quixotic to aspire for a teacher able
to provide with abstract floating point number slack values. Perhaps it is more
realistic to assume instead that the teacher can provide with some rich,
high-level representation useful to estimate the sought-after slack values.  This
reasoning crystallizes into the SVM+ objective function from \citep{Vapnik09}:
\begin{equation}\label{eq:similarity-control}
  L(w,w^\star,b,b^\star, \alpha, \beta) =
  \underbrace{\frac{1}{2}\|w\|^2 + \sum_{i=1}^n \alpha_i - \sum_{i=1}^n
  \alpha_i y_i f_i}_{\textrm{separable SVM objective}}
  + \underbrace{\frac{\gamma}{2} \|w^\star\|^2 + \sum_{i=1}^n (\alpha_i + \beta_i
  - C) {f^\star_i}}_{\textrm{corrections from teacher}},
\end{equation}
where $f_i := \langle w, x_i \rangle +b$ is the decision boundary at $x_i$, and
$f^\star_i := \langle w^\star, x^\star_i \rangle +b^\star$ is the teacher
correcting function at the same location. The SVM+ objective function matches
the objective function of non-separable SVM when we replace the correcting
functions $f^\star_i$ with the slacks $\xi_i$. Thus, skilled teachers provide
with privileged information $x^\star_i$ highly informative about the slack values
$\xi_i$. Such privileged information allows for simple correcting functions
$f^\star_i$, and the easy estimation of these correcting functions is a proxy
to $O(n^{-1})$ fast learning rates. Technically, this amounts to
saying that a teacher is helpful whenever the capacity of her correcting
functions is much smaller than the capacity of the student decision
boundary.

In \emph{knowledge transfer} \citep{VapIzm15} the teacher fits a function
$f_t(x^\star) = \sum_{j=1}^m \alpha^\star_j k^\star(u^\star_j,x^\star)$ on the
input-output pairs $\{(x^\star_i,y_i)\}_{i=1}^n$ and $f_t \in \mathcal{F}_t$,
to find the best reduced set of prototype or basis points
$\{u^\star_j\}_{j=1}^m$.  Second, the student fits one function $g_j$ per set
of input-output pairs $\{(x_i, k^\star(u^\star_j,x^\star_i))\}_{i=1}^n$, for
all $1 \leq j \leq m$. Third, the student fits a new vector of coefficients
$\alpha \in \mathbb{R}^m$ to obtain the final student function $f_s(x) =
\sum_{j=1}^m \alpha_j g_j(x)$, using the input-output pairs
$\{(x_i,y_i)\}_{i=1}^n$ and $f_s \in \mathcal{F}_s$.  Since the representation
$x^\star_i$ is intelligent, we assume that the function class $\mathcal{F}_t$
has small capacity, and thus allows for accurate estimation under small
sample sizes.

Distillation differs from similarity control in three ways. First, distillation 
is not restricted to SVMs.  Second, while the SVM+ solution contains twice the amount
of parameters than the original SVM, the user can choose a priori the amount of parameters in the distilled
classifier. Third, SVM+ learns the teacher
correcting function and the student decision boundary simultaneously,
but distillation proceeds sequentially: first with the teacher, then with the
student.  On the other hand, knowledge transfer is closer in spirit to
distillation, but the two techniques differ: while similarity control relies on a
student that purely imitates the hidden representation of a low-rank kernel
machine, distillation is a trade-off between imitating soft predictions and hard
labels, using arbitrary learning algorithms.

The framework of learning using privileged information enjoys theoretical
analysis \citep{Pechyony10a} and multiple applications, including ranking
\citep{Sharmanska13}, computer vision \citep{Sharmanska14,Lopez-Paz14},
clustering \citep{Feyereisl12}, metric learning \citep{Fouad13}, Gaussian
process classification \citep{Hernandez14}, and finance
\citep{ribeiro2010financial}. \citet{lapin2014learning} show
that learning using privileged information is a particular instance of
importance weighting.

\section{Generalized distillation}\label{sec:gendistillation}
We now have all the necessary background to describe \emph{generalized
distillation}. To this end, consider the data
$\{(x_i,x^\star_i,y_i)\}_{i=1}^n$. Then, the process of {generalized
distillation} is as follows:
\begin{enumerate}
  \item Learn teacher $f_t\in\mathcal{F}_t$ using the input-output pairs
  $\{(x^\star_i, y_i)\}_{i=1}^n$ and Eq.~\ref{eq:obj1}.
  \item Compute teacher soft labels $\{\sigma(f_t(x^\star_i)/T)\}_{i=1}^n$,
  using temperature parameter $T>0$.
  \item Learn student $f_s\in\mathcal{F}_s$ using the input-output pairs
  $\{(x_i, y_i)\}_{i=1}^n$, $\{(x_i, s_i)\}_{i=1}^n$, Eq. 
  \ref{eq:obj2}, and imitation parameter $\lambda \in [0,1]$.\footnote{Note
  that these three steps could be combined into a joint end-to-end optimization
  problem. For simplicity, our numerical simulations will take each of these
  three steps sequentially.} 
\end{enumerate}

We say that generalized distillation reduces to \emph{Hinton's distillation} if
$x^\star_i = x_i$ for all $1 \leq i \leq n$ and $|\mathcal{F}_s|_{\textrm{C}}
\ll |\mathcal{F}_t|_{\textrm{C}}$, where $|\cdot|_C$ is an appropriate function
class capacity measure. Conversely, we say that generalized distillation
reduces to \emph{Vapnik's learning using privileged information} if $x^\star_i$
is a privileged description of $x_i$, and $|\mathcal{F}_s|_{\textrm{C}} \gg
|\mathcal{F}_t|_{\textrm{C}}$.

This comparison reveals a subtle difference between Hinton's distillation and
Vapnik's privileged information.  In Hinton's distillation, $\mathcal{F}_t$ is
\emph{flexible}, for the teacher to exploit her \emph{general purpose}
representation $x^\star_i = x_i$ to learn intricate patterns from \emph{large}
amounts of labeled data. In Vapnik's privileged information, $\mathcal{F}_t$ is
\emph{simple}, for the teacher to exploit her \emph{rich} representation
$x^\star_i \neq x_i$ to learn intricate patterns from \emph{small} amounts of
labeled data.  The space of privileged information is thus a specialized space,
one of ``metaphoric language''. In our running example of biopsy images, the
space of medical reports is much more specialized than the space of pixels,
since the space of pixels can also describe buildings, animals, and other
unrelated concepts.  In any case, the teacher must develop a language that
effectively communicates information to help the student come up with better
representations. The teacher may do so by incorporating invariances, or biasing
them towards being robust with respect to the kind of distribution shifts that
the teacher may expect at test time.  In general, having a teacher is one
opportunity to learn characteristics about the decision boundary which are not
contained in the training sample, in analogy to a good Bayesian prior.

\subsection{Why does generalized distillation work?}
Recall our three actors: the student function $f_{s} \in \mathcal{F}_s$, the
teacher function $f_{t} \in \mathcal{F}_t$, and the real target function of
interest to both the student and the teacher, $f \in \mathcal{F}$. For simplicity, consider \emph{pure
distillation} (set the imitation parameter to $\lambda = 1$).
Furthermore, we will place some assumptions about how the student, teacher, and
true function interplay when learning from $n$ data.  First, assume that the
student may learn the true function at a slow rate
\begin{equation*}
  R(f_{s}) - R(f) \leq O\left(\frac{|\mathcal{F}_s|_\textrm{C}}{\sqrt{n}}\right) + \varepsilon_s,
\end{equation*}
where the $O(\cdot)$ term is the estimation error, and $\varepsilon_s$ is the
approximation error of the student function class $\mathcal{F}_s$ with respect
to $f \in \mathcal{F}$. Second, assume that the better representation of the
teacher allows her to learn at the fast rate 
\begin{equation*}
  R(f_{t}) - R(f) \leq O\left(\frac{|\mathcal{F}_t|_\textrm{C}}{n}\right) +
  \varepsilon_t,
\end{equation*}
where $\varepsilon_t$ is the approximation error of the teacher function class
$\mathcal{F}_t$ with respect to $f \in \mathcal{F}$. Finally, assume that when
the student learns from the teacher, she does so at the rate
\begin{equation*}
  R(f_{s}) - R(f_{t}) \leq
  O\left(\frac{|\mathcal{F}_s|_\textrm{C}}{n^{\alpha}}\right) + \varepsilon_l,
\end{equation*}
where $\varepsilon_l$ is the approximation error of the student function class
$\mathcal{F}_s$ with respect to $f_t \in \mathcal{F}_t$, and $\frac{1}{2} \leq
\alpha \leq 1$. Then, the rate at which the student learns the true function
$f$ admits the alternative expression 
\begin{align*}
  R(f_{s})-R(f) &= R(f_{s})-R(f_{t})+R(f_{t})-R(f)\\
                  &\leq
                  O\left(\frac{|\mathcal{F}_s|_\textrm{C}}{n^{\alpha}}\right)
                  + \varepsilon_l +
                  O\left(\frac{|\mathcal{F}_t|_\textrm{C}}{n}\right) +
                  \varepsilon_t\\
                  &\leq O\left(\frac{|\mathcal{F}_s|_\textrm{C} +
                  |\mathcal{F}_t|_\textrm{C}}{n^{\alpha}}\right) +
                  \varepsilon_l + \varepsilon_t,
\end{align*}
where the last inequality follows because $\alpha \leq 1$. Thus, the question
at hand is to argue, for a given learning problem, if the inequality
\begin{equation*}
  O\left(\frac{|\mathcal{F}_s|_\textrm{C} +
  |\mathcal{F}_t|_\textrm{C}}{n^{\alpha}}\right) + \varepsilon_l +
  \varepsilon_t \leq O\left(\frac{|\mathcal{F}_s|_\textrm{C}}{\sqrt{n}}\right)
  + \varepsilon_s
\end{equation*}
holds.  The inequality highlights that the benefits of learning with a teacher
arise due to i) the capacity of the teacher being small, ii) the approximation
error of the teacher being smaller than the approximation error of the student,
and iii) the coefficient $\alpha$ being greater than $\frac{1}{2}$.
Remarkably, these factors embody the assumptions of privileged information from
\citet{VapIzm15}. The inequality is also reasonable under the main assumption
in \citep{Hinton15}, which is $\varepsilon_s \gg \varepsilon_t +
\varepsilon_l$.  Moreover, the inequality highlights that the teacher is most
helpful in low data regimes, such as small datasets, Bayesian optimization,
reinforcement learning, domain adaptation, transfer learning, or in the initial
stages of online and reinforcement learning.

We believe that the ``$\alpha > \frac{1}{2}$ case'' is a general situation,
since soft labels (dense vectors with a real number of information per class)
contain more information than hard labels (one-hot-encoding vectors with one
bit of information per class) per example, and should allow for faster
learning. This additional information, also understood as label uncertainty, 
relates to the acceleration in SVM+ due to the knowledge of slack values. Since a
good teacher smoothes the decision boundary and instructs the student to fail
on difficult examples, the student can focus on the remaining body of data.
Although this translates into the unambitious ``whatever my
teacher could not do, I will not do'', the imitation parameter $\lambda \in [0,1]$
in \eqref{eq:obj2} allows to follow this rule safely, and fall back to regular
learning if necessary.

\subsection{Extensions}
\paragraph{Semi-supervised learning} We now extend generalized distillation to
the situation where examples lack regular features, privileged features,
labels, or a combination of the three. In the following, we denote missing
elements by $\square$. For instance, the example $(x_i, \square, y_i)$ has no
privileged features, and the example $(x_i,x^\star_i,\square)$ is missing its
label. Using this convention, we introduce the \emph{clean subset} notation
\begin{equation*}
  c(S) = \{ v : v \in S,  v_i \neq \square \,\, \forall i \}.
\end{equation*}
Then, semi-supervised generalized distillation walks the same three steps as
generalized distillation, enumerated at the beginning of
Section~\ref{sec:gendistillation}, but uses the appropriate clean subsets
instead of the whole data.  For example, the semi-supervised extension of
distillation allows the teacher to prepare soft labels for all
the unlabeled data $c(\{(x_i,x^\star_i)\}_{i=1}^n)$. These additional
soft-labels are additional information available to the student to learn the teacher
representation $f_t$.

\paragraph{Learning with the Universum} The unlabeled data
$c(\{x_i,x^\star_i\}_{i=1}^n)$ can belong to one of the classes of interest, or be
\emph{Universum} data \citep{Weston06,Chapelle07}. Universum data may have
labels: in this case, one can exploit these additional labels by i) training a
teacher that distinguishes amongst all classes (those of interest and those 
from the Universum), ii) computing soft class-probabilities only for the
classes of interest, and iii) distilling these soft probabilities into a
student function.

\paragraph{Learning from multiple tasks} Generalized distillation applies to
some domain adaptation, transfer learning, or multitask learning scenarios. On
the one hand, if the multiple tasks share the same labels $y_i$ but differ in
their input modalities, the input modalities from the source tasks are
privileged information.  On the other hand, if the multiple tasks share the
same input modalities $x_i$ but differ in their labels, the labels from the
source tasks are privileged information. In both cases, the regular student
representation is the input modality from the target task.

\paragraph{Curriculum and reinforcement learning} We conjecture that the
uncertainty in the teacher soft predictions can be used as a mechanism to rank
the difficulty of training examples, and use these ranks for curriculum
learning \citep{bengio2009curriculum}. Furthermore, distillation 
resembles imitation, a technique that learning agents could exploit in
\emph{reinforcement learning} environments.

\subsection{A causal perspective on generalized distillation}\label{sec:causal}

The assumption of \emph{independence of cause and mechanisms} states that ``the
probability distribution of a cause is often independent from the process
mapping this cause into its effects'' \citep{Scholkopf12}.  Under this
assumption, for instance, \emph{causal learning problems} ---i.e., those where
the features cause the labels--- do not benefit from semi-supervised learning,
since by the independence assumption, the marginal distribution of the features
contains no information about the function mapping features to labels.
Conversely, \emph{anticausal learning problems} ---those where the labels cause
the features--- may benefit from semi-supervised learning. 

Causal implications also arise in generalized distillation.  First, if the
privileged features $x^\star_i$ only add information about the marginal
distribution of the regular features $x_i$, the teacher should be able to help
only in anticausal learning problems. Second, if the teacher provides
additional information about the conditional distribution of the labels $y_i$
given the inputs $x_i$, it should also help in the causal setting. We will confirm this hypothesis in the next section.

\section{Numerical simulations}
We now present some experiments to illustrate when the
distillation of privileged information is effective, and when it is not.  The
necessary Python code to replicate all the following experiments is available
at \url{http://github.com/lopezpaz}.

We start with four synthetic experiments, designed to minimize modeling
assumptions and to illustrate different prototypical types of privileged
information. These are simulations of logistic regression models repeated over
$100$ random partitions, where we use $n_\text{tr} = 200$ samples for training,
and $n_\text{te} = 10,000$ samples for testing.  The dimensionality of the
regular features $x_i$ is $d=50$, and the involved separating hyperplanes
$\alpha \in \mathbb{R}^d$ follow the distribution $\mathcal{N}(0,I_d)$.  For each
experiment, we report the test accuracy when i) using the teacher explanations
$x^\star_i$ at both train and test time, ii) using the regular features $x_i$
at both train and test time, and iii) distilling the teacher explanations into
the student classifier with $\lambda = T = 1$. 

\paragraph{1. Clean labels as privileged information.} We sample triplets $(x_i,
x^\star_i, y_i)$ from:
\begin{align*}
  x_i       &\sim \mathcal{N}(0,I_d)\\
  x^\star_i &\leftarrow \langle \alpha, x_i \rangle\\
  \varepsilon_i &\sim \mathcal{N}(0,1)\\
  y_i       &\leftarrow \mathbb{I}((x^\star_i + \varepsilon_i) > 0).
\end{align*}
Here, each teacher explanation $x^\star_i$ is the exact distance to the
decision boundary for each $x_i$, but the data labels $y_i$ are corrupt. This
setup aligns with the assumptions about slacks in the similarity control
framework of \citet{Vapnik09}. We obtained a privileged test classification
accuracy of $96 \pm 0\%$, a regular test classification accuracy of $88\pm
1\%$, and a distilled test classification accuracy of $95\pm
1\%$. This illustrates that distillation of privileged information is an
effective mean to detect outliers in label space.

\paragraph{2. Clean features as privileged information} We sample 
triplets $(x_i, x^\star_i, y_i)$ from:
\begin{align*}
  x^\star_i     &\sim \mathcal{N}(0,I_d)\\
  \varepsilon_i &\sim \mathcal{N}(0,I_d)\\
  x_i     &\leftarrow x^\star_i + \varepsilon\\
  y_i       &\leftarrow \mathbb{I} \left(\langle \alpha, x^\star_i \rangle > 0\right).
\end{align*}
In this setup, the teacher explanations $x^\star_i$ are clean versions of the
regular features $x_i$ available at test time.  We obtained a privileged test
classification accuracy of $90 \pm 1\%$, a regular test classification accuracy
of $68\pm 1\%$, and a distilled test classification accuracy of
$70\pm 1\%$. This improvement is not statistically significant. This is because
the intelligent explanations $x^\star_i$ are independent from the noise
$\varepsilon_i$ polluting the regular features $x_i$.  Therefore, there exists
no additional information transferable from the teacher to the student.

\paragraph{3. Relevant features as privileged information} We sample triplets
$(x_i, x^\star_i, y_i)$
from:
\begin{align*}
  x_i       &\sim \mathcal{N}(0,I_d)\\
  x^\star_i &\leftarrow x_{i,J}\\
  y_i       &\leftarrow \mathbb{I}(\langle \alpha_J, x^\star_i\rangle > 0),
\end{align*}
where the set $J$, with $|J| = 3$, is a subset of the variable indices $\{1,
\ldots, d\}$ chosen at random but common for all samples. In another words, the
teacher explanations indicate the values of the variables relevant for
classification, which translates into a reduction of the
dimensionality of the data that we have to learn from.  We obtained a
privileged test classification accuracy of $98 \pm 0\%$, a regular test
classification accuracy of $89\pm 1\%$, and a distilled test
classification accuracy of $97\pm 1\%$. This illustrates that distillation on
privileged information is an effective tool for feature selection.

\paragraph{4. Sample-dependent relevant features as privileged information}
Sample triplets
\begin{align*}
  x_i       &\sim \mathcal{N}(0,I_d)\\
  x^\star_i &\leftarrow x_{i,{J_i}}\\
  y_i       &\leftarrow \mathbb{I}(\langle \alpha_{J_i}, x^\star_i\rangle > 0),
\end{align*}
where the sets $J_i$, with $|J_i| = 3$ for all $i$, are a subset of the
variable indices $\{1, \ldots, d\}$ chosen at random for each sample
$x^\star_i$. One interpretation of such model is the one of bounding boxes in
computer vision: each high-dimensional vector $x_i$ would be an image, and each
teacher explanation $x^\star_i$ would be the pixels inside a bounding box
locating the concept of interest \citep{Sharmanska13}.
We obtained a privileged test classification accuracy of $96 \pm 2\%$, a
regular test classification accuracy of $55\pm 3\%$, and a distilled
test classification accuracy of $0.56\pm 4\%$. Note that
although the classification is linear in $x^\star$, this is not the case in
terms of $x$. Therefore, although we have misspecified the function class
$\mathcal{F}_s$ for this problem, the distillation approach did not deteriorate
the final performance.

The previous four experiments set up causal learning problems. In the second
experiment, the privileged features $x^\star_i$ add no information about the
target function mapping the regular features to the labels, so the causal
hypothesis from Section~\ref{sec:causal} justifies the lack of improvement. The
first and third experiments provide privileged information that adds
information about the target function, and therefore is beneficial to distill
this information. The fourth example illustrates that 
the privileged features adding information about
the target function is not a sufficient condition for improvement.

\begin{figure}
  \begin{subfigure}{0.5\textwidth}
    \begin{center}
    \includegraphics[width=0.85\textwidth]{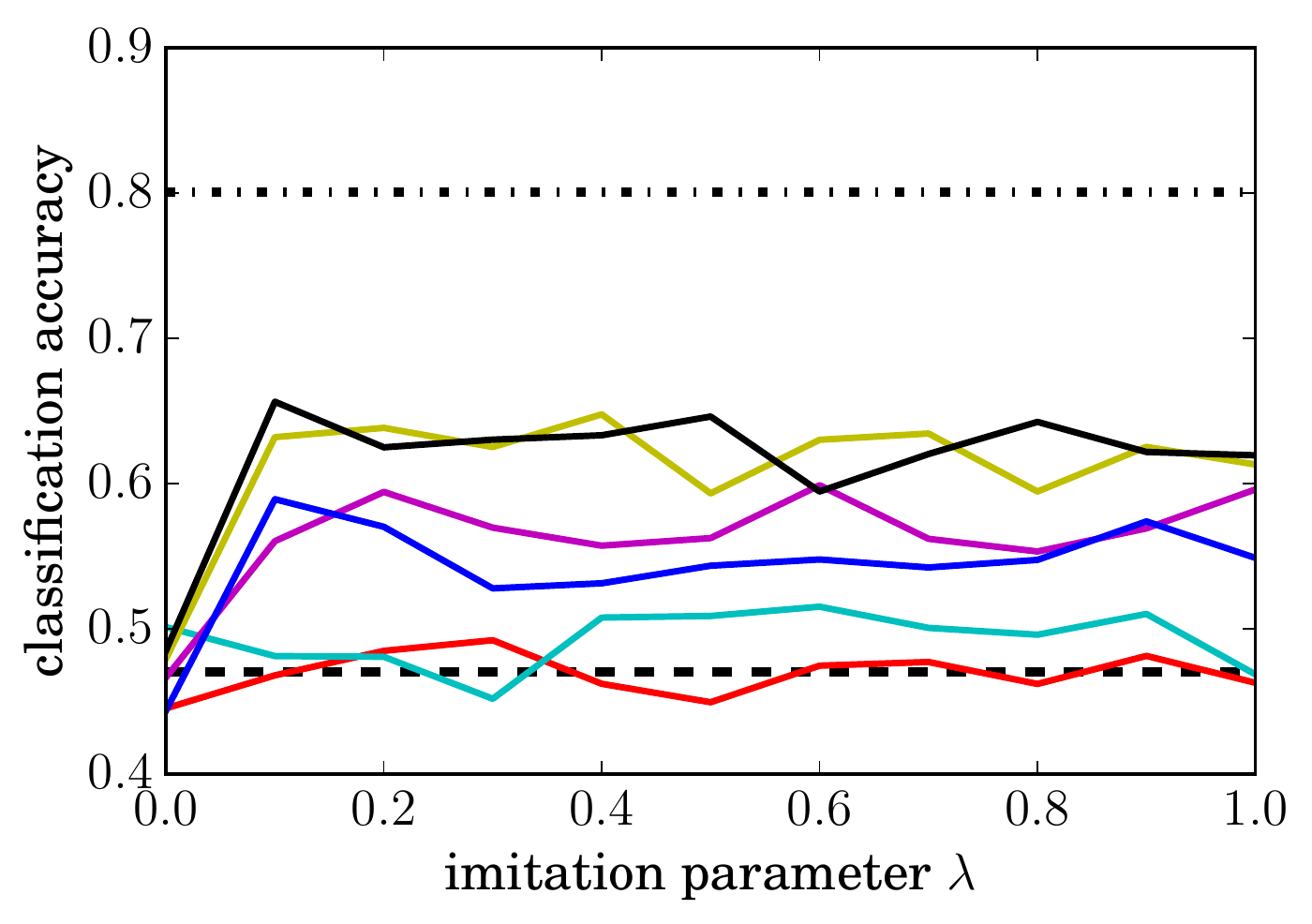}
    \end{center}
  \end{subfigure}
  \begin{subfigure}{0.5\textwidth}
    \begin{center}
    \includegraphics[width=0.85\textwidth]{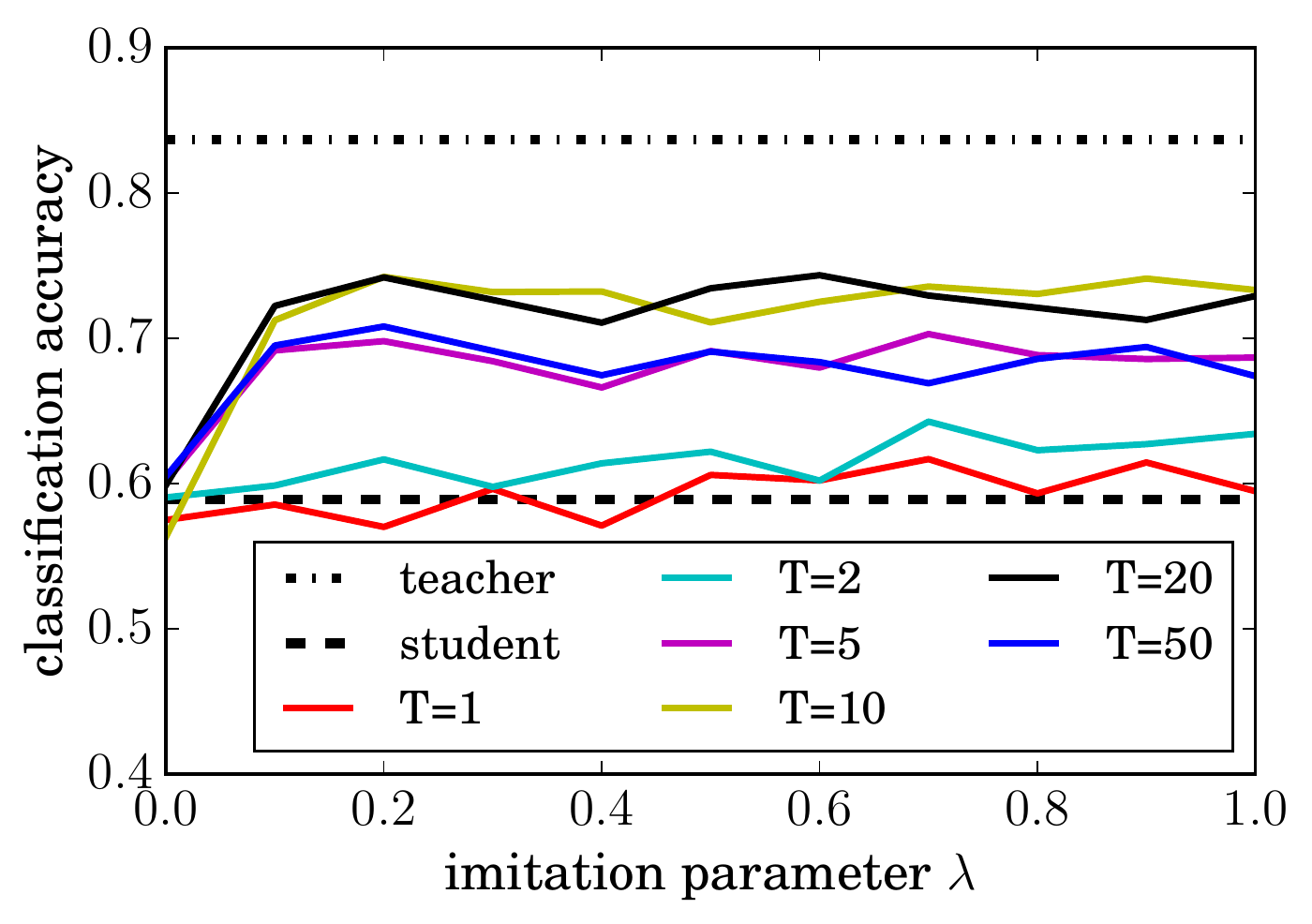}
    \end{center}
  \end{subfigure}
  \caption{Results on MNIST for 300 samples (left) and 500 samples (right).}
  \label{fig:mnist}
\end{figure}

\paragraph{5. MNIST handwritten digit image classification} The privileged
features are the original 28x28 pixels MNIST handwritten digit images
\citep{MNIST}, and the regular features are the same images downscaled to 7x7
pixels. We use $300$ or $500$ samples to train both the teacher and the
student, and test their accuracies at multiple levels of temperature and
imitation on the full test set. Both student and teacher are neural networks of
composed by two hidden layers of $20$ rectifier linear units and a softmax
output layer (the same networks are used in the remaining experiments).
Figure~\ref{fig:mnist} summarizes the results of this experiment, where we see
a significant improvement in classification accuracy when distilling the
privileged information, with respect to using the regular features alone. As
expected, the benefits of distillation diminished as we further increased the
sample size.

\begin{figure}
  \begin{subfigure}{0.5\textwidth}
    \begin{center}
    \includegraphics[width=.85\textwidth]{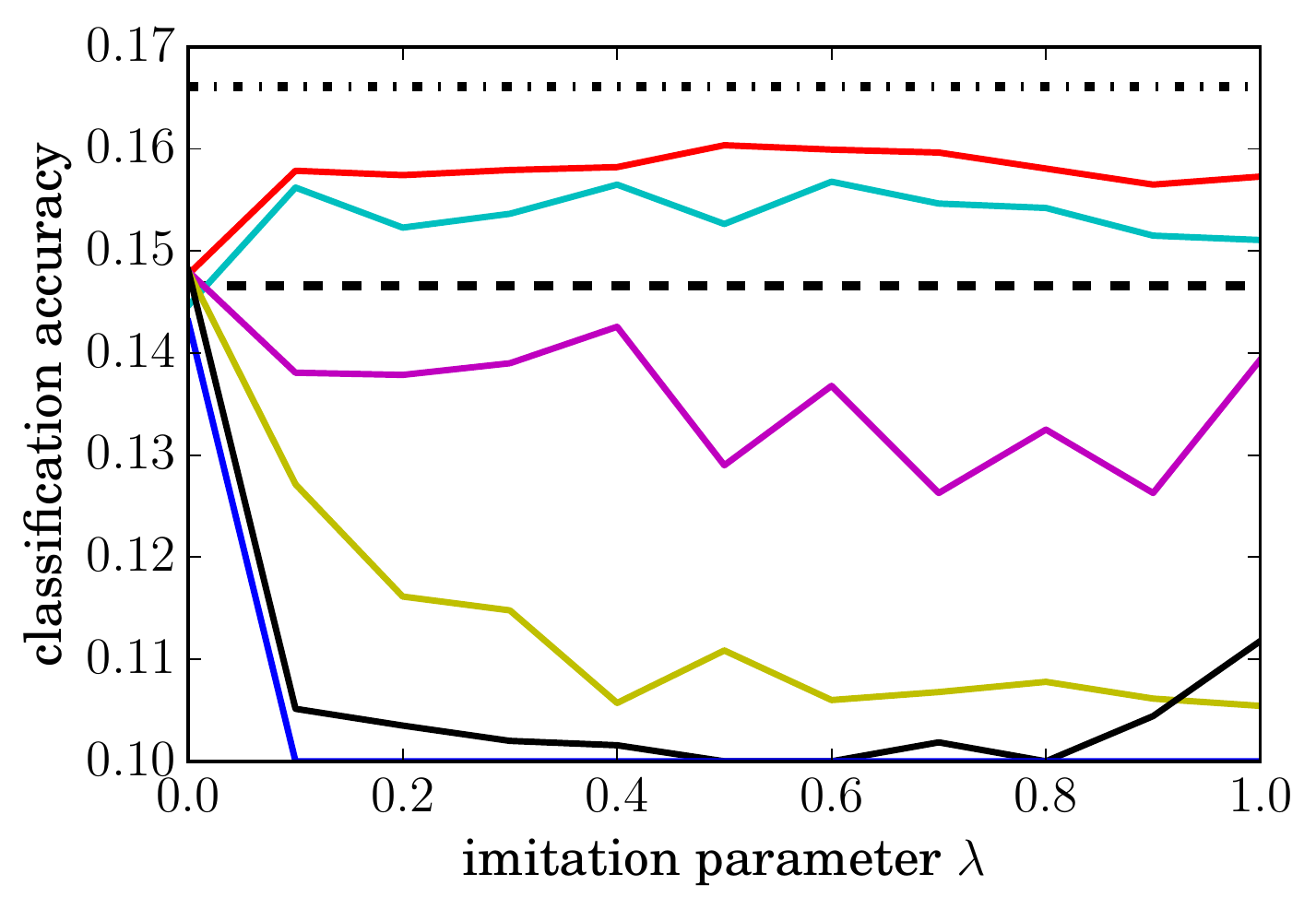}
    \end{center}
  \end{subfigure}
  \begin{subfigure}{0.5\textwidth}
    \begin{center}
    \includegraphics[width=.85\textwidth]{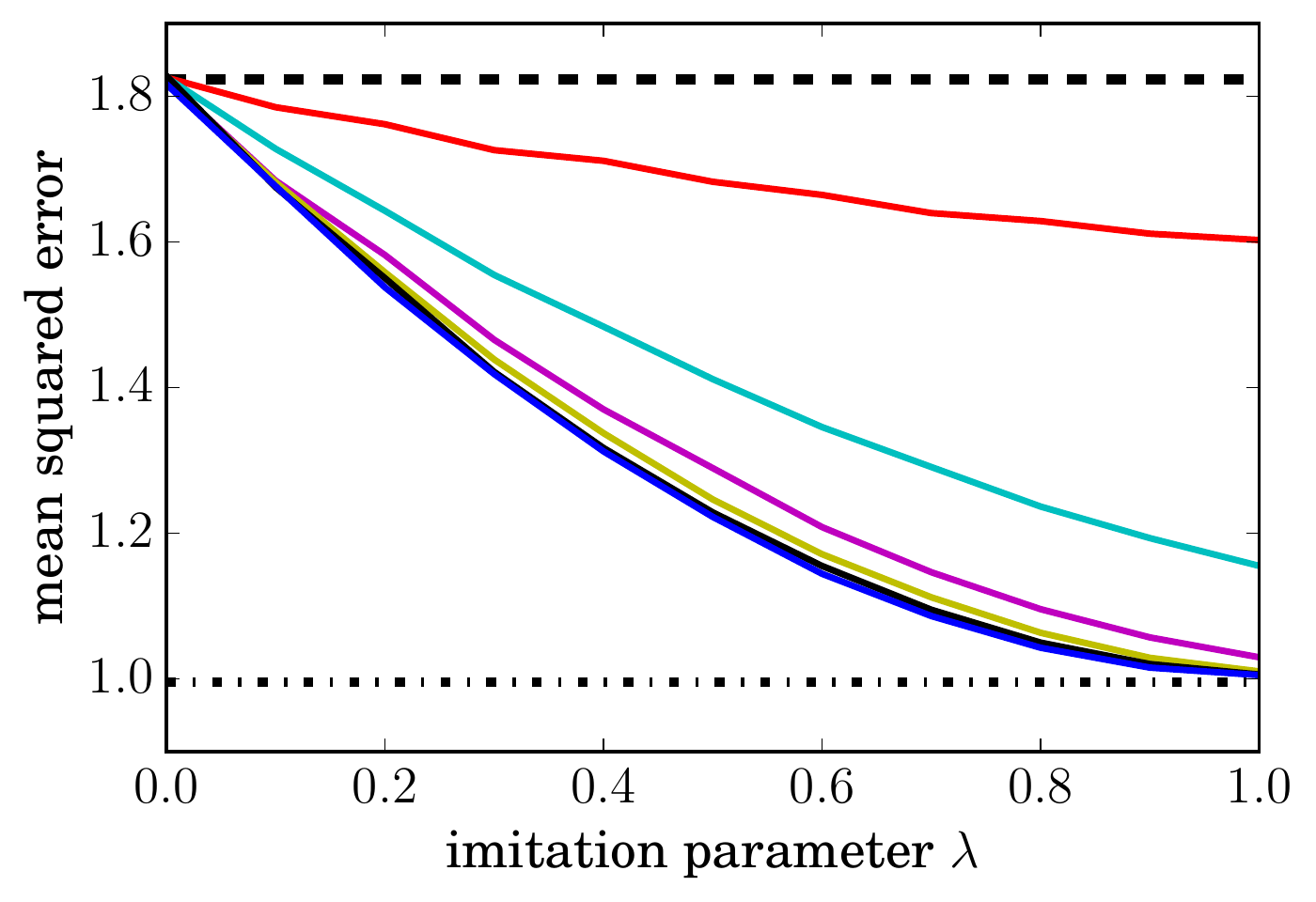}
    \end{center}
  \end{subfigure}
  \caption{Results on CIFAR 10 (left) and SARCOS (right).}
  \label{fig:others}
\end{figure}

\paragraph{6. Semisupervised learning} We explore the semisupervised capabilities
of generalized distillation on the CIFAR10 dataset \citep{CIFAR10}. Here, the
privileged features are the original 32x32 pixels CIFAR10 color images, and the
regular features are the same images when polluted with additive Gaussian
noise. We provide labels for $300$ images, and unlabeled privileged and regular
features for the rest of the training set. Thus, the teacher trains on $300$
images, but computes the soft labels for the whole training set of $50,000$
images. The student then learns by distilling the $300$ original hard labels
and the $50,000$ soft predictions. As seen in Figure~\ref{fig:others}, the soft
labeling of unlabeled data results in a significant improvement with respect to
pure student supervised classification. Distillation on the $300$ labeled
samples did not improve the student performance. This illustrates the
importance of semisupervised distillation in this data.  We believe that the
drops in performance for some distillation temperatures are due to the lack of
a proper weighting between labeled and unlabeled data in \eqref{eq:obj2}.

\paragraph{7. Multitask learning} The SARCOS
dataset \citep{SARCOS} characterizes the 7 joint torques of a robotic arm given
21 real-valued features.  Thus, this is a multitask learning problem, formed by
7 regression tasks. We learn a teacher on $300$ samples to predict each of the
7 torques given the other 6, and then distill this knowledge into a student who
uses as her regular input space the 21 real-valued features.
Figure~\ref{fig:others} illustrates the performance improvement in mean squared
error when using generalized distillation to address the multitask learning
problem. When distilling at the proper temperature, distillation allowed
the student to match her teacher performance.  

\clearpage
\newpage
\subsection*{Acknowledgments}
We thank discussions with R. 
Nishihara, R. Izmailov, I. Tolstikhin, and C. J. Simon-Gabriel.

\bibliographystyle{iclr2016_conference}
\bibliography{distillation}
\end{document}